# Cross Modal Data Discovery over Structured and Unstructured Data Lakes

Mohamed Y. Eltabakh[+], Mayuresh Kunjir[★], Ahmed Elmagarmid[+], Mohammad Shahmeer Ahmad[+]
[+]*Qatar Computing Research Institute, HBKU, Qatar*   [★]*Amazon Web Services, Germany*[*]
meltabakh@hbku.edu.qa, mkunjir@amazon.de, aelmagarmid@hbku.edu.qa, mohammadshahmeerah@hbku.edu.qa

## ABSTRACT

Organizations are collecting increasingly large amounts of data for data-driven decision making. These data are often dumped into a centralized repository, e.g., a data lake, consisting of thousands of structured and unstructured datasets. Perversely, such mixture of datasets makes the problem of *discovering* elements (e.g., tables or documents) that are relevant to a user's query or an analytical task very challenging. Despite the recent efforts in *data discovery*, the problem remains widely open especially in the two fronts of (1) discovering relationships and relatedness across structured and unstructured datasets–where existing techniques suffer from either scalability, being customized for a specific problem type (e.g., entity matching or data integration), or demolishing the structural properties on its way, and (2) developing a holistic system for integrating various similarity measurements and sketches in an effective way to boost the discovery accuracy.

In this paper, we propose a new data discovery system, named CMDL, for addressing these two limitations. CMDL supports the data discovery process over both structured and unstructured data while retaining the structural properties of tables. As a result, CMDL is the only system to date that empowers end-users to seamlessly pipeline the discovery tasks across the two modalities. We propose a novel multi-modal embedding representation that captures the similarities between text documents and tabular columns. The model training relies on labeled datasets generated though *weak supervision*, and thus the system is domain agnostic and easily generalizable. We evaluate CMDL on three real-world data lakes with diverse applications and show that our system is significantly more effective for cross-modality discovery compared to the search-based baseline techniques. Moreover, CMDL is more accurate and robust to different data types and distributions compared to the state-of-the-art systems that are limited to only the structured datasets.

## 1 INTRODUCTION

The democratization of data science has resulted in enterprises drowning in their own collected data. Large enterprises routinely have a huge collection of structured and unstructured data in their data repositories, referred to as *data lakes*. Performing *data discovery* where the goal is to find relevant elements (e.g., tables or documents) for a given user's query or an analytical task from such massive and unorganized data lakes is a major challenge faced by data scientists [3, 26, 48, 51, 64]. Data discovery is one of the initial steps of data preparation that plays an important role in various downstream data processing tasks including data exploration, data integration, and data enrichment [37, 40, 42, 63, 71].

Data discovery is emerging as a practical and core problem to modern enterprises as exemplified by the recent interest from both academia and industry [1, 2, 15, 31, 38, 49]. Nevertheless, it remains a challenging problem due to several factors. First, data is collected and published by disparate sources, and thus do not conform to a common format or structure. A vast amount of this data is in the form of unstructured text documents, e.g., research articles, social media posts, emails, news reports, etc. [17, 28]. As a result, both data modalities (structured and unstructured) are equally critical, and any system that overlooks either of them is lame. The second factor is the lack of consistent and sufficient data description and metadata. Therefore, a robust data discovery system must reason about and establish relationships of relatedness using multiple and diverse similarity signals, where each could be possibly weak. Lastly, a data discovery system needs to scale well to very large volumes of data as more organizations and governments are publishing more data for transparency [62].

The majority of existing work limits the discovery process to only the structured portion of data lakes, e.g., [15, 31, 33, 34, 67]. And thus, these systems do not extend to unstructured data and hit the wall of the first challenge mentioned above. In contrast, there are recent efforts trying to extend the discovery process across both structured and unstructured data sets, however they inherit several limitations. For example, some of these systems shift the focus entirely to the unstructured domain by transforming the structured data to unstructured documents [17, 19, 28]. This way, they demolish the structural properties and cut the way towards having a holistic discovery system across both modalities. Other systems move the two modalities to a third common space where everything is encoded as a set of subject-predicate-object triplets, e.g., [32]. Again, they lose the ability for supporting discoveries such as table joinability and unioinability in addition to encountering high overheads that hinder its scalability.

The goal of this work is to develop a more holistic data discovery system that can: (1) bring large unstructured data into the framework of data discovery, (2) enable analysts to intermix their structured and unstructured discovery tasks in a single pipeline seamlessly, and (3) be robust through novel integration of various content-based and metadata-based similarity signals–this is especially required because unstructured data inherit very different characteristics compared to structured tables.

**Motivation Example (Pharmaceutical Usecase):** *Pharmaceutical databases such as DrugBank consist of an exhaustive listing of drugs, enzymes, and genes along with their interactions with other drugs or related proteins [4]. Domain experts periodically curate such databases with findings from research literature, MedLine reports [8], FDA MedWatch reports [7], and other sources. Such data are then used for accurate drug-drug interaction prediction, drug safety surveillance, and drug design optimizations. Assume our data lake encompasses* all these rich data sets, and the analysts are exploring the data in the wild, i.e., without any prior knowledge of the available data sources. Now, an analyst studying enzyme *"Thymidylate Synthase"* would like to iteratively explore the data lake to discover relevant information (as illustrated in Figure 1). For example, she can be interested at first in finding documents related to this enzyme (Q1 in the figure). The system returns such documents as depicted in the left side of the figure. Then, she poses the second question (Q2) trying to find curated tables in the database related to the 1st document (or sub-document by highlighting sentences of interest). The system should provide table *"Enzyme_Targets"* as an answer since both (the table and the document talk about enzymes). Then, progressing forward with her exploration, she finds another returned document that talks about drug *"Pemetrexed"* and its relation to enzyme *"Thymidylate Synthase"*. As such, she poses Q3 to potentially find more tables in the database supporting this relationship. The system should provide the two tables *"Enzyme_Targets"* and *"Drugs"* as high likelihood candidates to Q3. One step further, the analyst would like to know if there are more relationships in the database around this drug. And thus, she raises Q4 and the system provides two potentially joinable tables (i.e., *"Enzyme_Targets"* and *"Drug_Interactions"*). The former table seems central to this exploration as it has been reached from multiple paths, and hence the analyst issues Q5 to find more related and unionable tables.

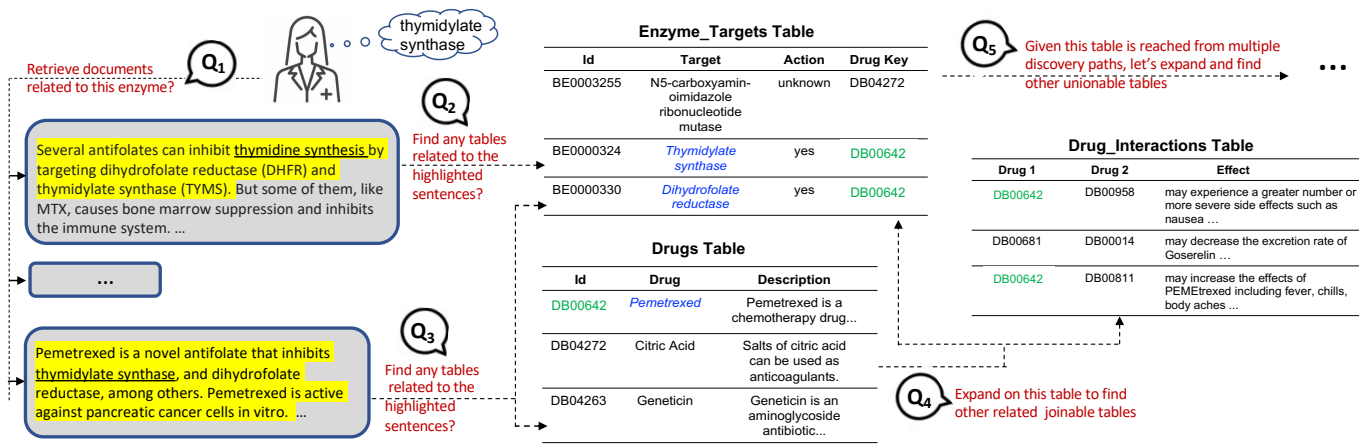

Figure 1: An example data discovery scenario on a lake of structured and unstructured pharmaceutical data.

Evidently, the kind of relationships, similarity measures, and sketches needed to be maintained and exploited to answer Question Q1 (within the unstructured data modality) are very different from those needed to answer Questions Q2 and Q3 (across modalities), and also very distinct from those needed to answer Question Q4 (within the structured data modality). To the best of our knowledge, there is no single holistic system that can support such data discovery chain as highlighted above.

**Solution overview:** We propose a data discovery system, called CMDL[1], that enables the data discovery tasks demonstrated in the motivation example over a data lake of structured and unstructured data sets. To support cross-modal questions such as Q2 and Q3 (from Figure 1), we introduce a novel embedding-based joint representation that brings both unstructured documents and relational tables, more specifically the individual columns in each table, into a common vector space. We propose several optimizations over the model's loss function to achieve scalability and avoid biases as will be explained later. The model that creates these joint representations relies on supervised learning and requires the presence of a labeled training data set–which may not exist in the first place. Therefore, CMDL is equipped with a weakly-supervised training process that integrates multiple similarity signals over documents and columns to construct a labeled training data set. As such CMDL does not require a pre-existing labeled datasets, and hence, easily generalizable to different applications.

To support other types of data discovery questions such as Q1, Q4, and Q5 (from Figure 1), CMDL utilizes several types of signatures and sketches on both the actual data content and the contextual metadata. For example, for structured data, schema information and tables names are types of metadata. Similarly, for unstructured documents, the documents' names and their sources are the metadata. Then, we build appropriate indexes using state-of-the-art techniques on them, e.g., BM25 elastic search index [58] for keyword search and Locality Sensitive Hashing (LSH) index [69] for the set containment operations. While prior data discovery solutions are also based on similar sketches and indexes, e.g., [15, 19, 31, 33, 34], CMDL goes one step beyond by first leveraging them as similarity signals for the weak-supervised training process mentioned above, and second integrating these complementary elements into a single unified discovery system.

The various relationships discovered by the system form an Enterprise Knowledge Graph (EKG) on which a comprehensive query interface is developed. It extends the *SRQL* query engine from Aurum [31] with APIs and retrieval methods specific to CMDL, e.g., for supporting cross-modality discovery tasks.

In summary, we make the following contributions:

(1) Introducing a holistic data discovery system that treats both structured tabular data and unstructured documents as first-class citizens. Within a single discovery pipeline, end-users can seamlessly formulate discovery questions ranging from simple keyword search over either modality to complex discoveries spanning both modalities.
(2) Proposing a novel embedding-based joint representation for multi-modal data sets. The model training relies on weak supervision by combining various types of similarity signals. As such, the system can be adopted even under the absence of labeled data.
(3) Integrating existing sketches with a few alterations to improve the discovery accuracy even for the traditional joinability and unionability tasks by up to 30%.
(4) Contributing benchmarks to the research community for measuring the effectiveness and efficiency of data discovery solutions over structured and unstructured data.
(5) Evaluating CMDL on three real-world data lakes ranging from drug discovery to ML data augmentation. CMDL is not only more effective in discovering the complex relationships across the unstructured documents and the structured tables, but also more robust for the traditional discovery tasks (e.g., joinability and unionability) over the structured data compared to the state-of-the-art systems.

---

[1]CMDL–<u>C</u>ross <u>M</u>odal <u>D</u>ata <u>D</u>iscovery over Structured and Unstructured Data <u>L</u>akes–is pronounced as kam·dl.

## 2 CMDL OVERVIEW

In order to support data discovery pipelines similar to that presented in Figure 1, we need four building block elements. First, the notion of *discoverable elements* (*DEs*) that are the output from the discovery queries. Second, semantics for identifying the relationship between the *DEs*. Third, algorithms and data structures to efficiently discover and store the *DEs* and their relationships. Finally, an expressive query language for expressing the discovery tasks.

In this section, we first introduce the basic definitions of DEs and the target relationships, and then present the system's overall architecture that connects all components together. In the subsequent sections, we will discuss each component in detail.

### 2.1 Discoverable Elements and Relationships

**CMDL Discoverable Elements.** We borrow the concept of Discoverable Elements from Aurum [31], which is the abstract unit of data discovery. In CMDL, we treat each **column** and each **document** as the basic units of discovery for the structured and unstructured data sets, respectively. Based on columns, each **table** in the data lake is also treated as a higher-order DE. The intuition behind the column-level granularity is multifold. (1) Working at this granularity allows the system to scale well to massive datasets, (2) The column-level discovery is already the basic unit of discovery over structured data (e.g., joinability and unionability) [15, 35, 67], and thus it would work seamlessly in the entire CMDL framework, and (3) Columns typically have coherent semantics, e.g., drug names or city names, that embeddings can easily capture (unlike tuples which could have many attributes with very diverse semantics).

For simplicity, we assume each unstructured document is short, i.e., several sentences as illustrated in Figure 1. From a use case and end-user point of view, this assumption is realistic, because otherwise starting a discovery search with a large document, e.g., several pages, is very un-focused and will most probably lead to countless relationships of no interest. Moreover, from the system's backend point of view, this assumption is not restrictive because large documents can still be uploaded as physical units, however, CMDL will (logically) break each document into smaller DE units, e.g., paragraphs, for the discovery purpose.

**CMDL DE Relationship Types.**[2] In this paper, we emphasize three key relationship types between DEs as motivated by our example in Section 1.

*Doc$_{to}$Table (From Document to Tables).* A Table $T$ with column set $\mathbf{A}$ is related to a text document $D$ iff $\exists A_i \in \mathbf{A}$ s.t. $D$ and $A_i$ are related via overlapping values, semantic similarity, or metadata similarity, each with a relatedness score. The combined scores of the links $D \rightarrow A_i$ represent the strength of the relationship.

*Table$_J$Table (Joinable Tables).* Table $T$ with column set $\mathbf{A}$ is joinable to Table $T'$ with column set $\mathbf{A}'$ iff $\exists A_i \in \mathbf{A}$ and some $A'_j \in \mathbf{A}'$ s.t.: (a) $A_i$ and $A'_j$ have value overlap suggesting syntactic join, or (b) $A_i$ and $A'_j$ have semantic overlap suggesting semantic join.

*Table$_U$Table (Unionable Tables).* Table $T$ with column set $\mathbf{A}$ is unionable to Table $T'$ with column set $\mathbf{A}'$ iff a one-to-one mapping $H : \mathbf{A} \Rightarrow \mathbf{A}'$ exists wherein $\exists h \in H$ s.t. the column pair given by $h$ exhibits name, value, or semantic similarity. The combined similarity score of the column mapping gives the strength of the relation.

A relationship is said to exist from $D_i$ to $D_j$ if either their relationship strength exceeds a given threshold $\epsilon$, or alternatively, $D_j$ is within the top $k$ related DEs to $D_i$.

### 2.2 CMDL Architecture

Figure 2 illustrates the overall system's architecture. From left to right, both the text documents and structured tables go through a preprocessing phase. More specifically, each text document goes through an NLP-based pipeline for transformation, which ultimately results in converting a document into a column format consisting of a bag of words. In contrast, each column in the tabular data goes through heuristic-based tagging that labels the column based on the potential relationships and discovery tasks it will participate in. The next step is the *profiler*, which creates various types of sketches and statistics to be leveraged later in the discovery tasks for capturing syntactic and semantic relationships. These sketches then go to the *indexing framework* for building the appropriate indexes according to the individual sketch type (e.g., for metadata and content data, an elastic search index is constructed, whereas for min hash statistics an LSHEnsemble index is constructed).

The sketches created so far by the *profiler* are independent of and agnostic to any relationships between the text documents and the tabular columns. The *Multi-Modal Joint Representation* module is responsible for creating joint sketches that encode such relationships, i.e., related document-column pairs should have similar representations whereas unrelated pairs should have dissimilar representations in the new joint space. As will be explained in Section 4, this module consists of two sub-modules. The first one is the *Training Dataset Generator* that creates a training data set though weak supervision by leveraging various signals from the existing indexes. The second sub-module is the *Joint Representation Learning* that constructs a model for generating the desired joint representations. As depicted in Figure 2, this new representation is also passed to the *indexing framework* for constructing the appropriate indexes. The final module is the *EKG Builder*, which integrates the different types of relationships among all DE pairs into an Enterprise Knowledge Graph (EKG) along with its query interfaces and APIs.

---
[2]Simple relationships, e.g., finding documents or tabular columns containing certain keywords, are supported by the system but not explicitly highlighted here.

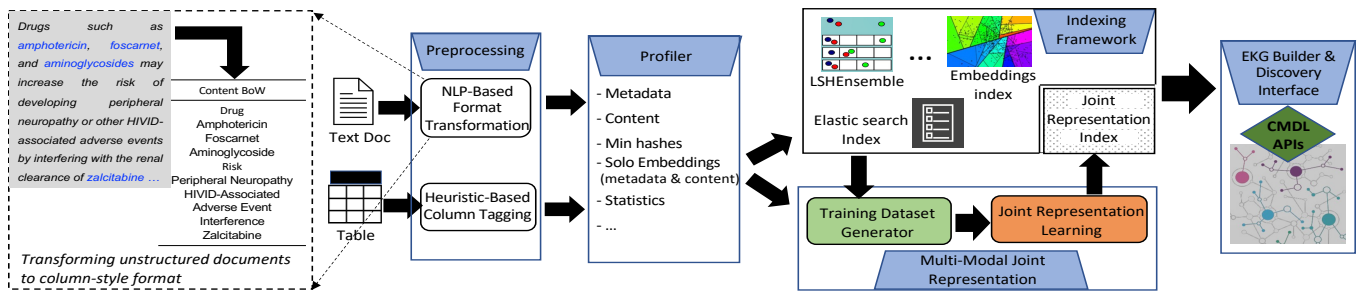

Figure 2: The CMDL system architecture.

## 3 SKETCHES AND INDEXING FRAMEWORK

In this section, we present the details of the preprocessing and profiler phases along with the indexing of the profiler-generated sketches.

**Documents Format Transformation.** In order to facilitate the discovery of cross-modality relationships, we opt for transforming unstructured documents using an NLP-based pipeline into column-style format. As such, all subsequent phases (including the *profiler*) will work on a unified column-style format. For that, CMDL adopts the Bag of Words technique [66] since it is a generic solution and does not require external knowledge, which suits our system well. Each document goes through the NLP phases of tokenization, stop-word removal, part-of-speech filtering (to retain only the noun terms), and lemmatization. We also filter out the words that occur in a large number of documents since they are non-discriminative. An example output of this process is depicted in the left side of Figure 2.

**Tabular Columns Tagging.** CMDL applies few heuristics to tag columns according to which discovery task(s) they can potentially participate in. Based on these tags, the *profiler* will later create certain types of sketches. For example, for document-column and keyword search discoveries, we filter out columns that are not text (e.g., numeric and date types), and categorical columns having few distinct values (e.g., below a certain percentage of the table's cardinally). In contrast, for PK-FK discoveries, we filter out date and long text columns.

**Profiler and Sketches Generation.** As highlighted in Section 1, different types of discovery questions would require different types of sketches to support them. The *profiler*'s task is to create various types of these sketches including:

• *Syntactic Similarity via Jaccard Distances.* A syntactic similarity between a pair of DEs can be measured by the percentage of their overlapping values. Several systems, e.g., [19, 31, 33], use **Jaccard Similarity** as a measure, where the similarity between two DEs (say sets $A$ and $B$) is computed as $|A \cap B|/|A \cup B|$. However, it is reported that this metric suffers poor performance if the domain sizes of $A$ and $B$ are very different [69]. Although this might not be an issue for discoveries within the tabular datasets of comparable sizes, it is an issue for cross-modality discoveries (e.g., Q2 and Q3 in Figure 1). Therefore, in CMDL, we adopt the **Jaccard Set Containment** score, which is an asymmetric metric from set $A$ (i.e., document side) to set $B$ (column side) and is measured as $|A \cap B|/|A|$. Nevertheless, for joinable and unionable discoveries (across two columns), the score is computed in both directions. To approximate the set containment measure, we build minwise hashing sketches as proposed in [69].

• *Semantic Similarity via Solo Embeddings.* One way for measuring semantic similarity between a pair of DEs is their proximity in a vector embedding space. Given that all inputs to the profiler (documents and columns) are represented as a collection of words, the profiler applies a pre-trained word embedding model, e.g., the fasttext model [16], on each word to generate its vector representation. And then, these word representations are aggregated using mean pooling [43] to construct a summarized representation at the column level.[3] We refer to these embedding vectors as **solo embeddings** since they are learned independently for each DE–unlike the joint embedding representation presented later.

• *Other Profiled Information.* The profiler also maintains additional information on each DE that will be used in different discovery tasks. This includes the metadata information–which is the column and table names for tabular columns and the titles for documents. The metadata is used to build LSH-based name and schema similarities across tables, and also used as part of the learned joint representation (Section 4). Moreover, for numeric columns, additional statistics are maintained, e.g., number of distinct values, domain size, min and max values. These statistics are used for building numeric-based overlap similarity as in [15, 31]

**Indexing Profiler-Generated Sketches.** For efficient search later on, each of the generated sketch types is indexed using an appropriate index structure. For example, solo embeddings are indexed using Annoy space partitioning structures [45] and minwise hashes are indexed using Locality Sensitive Hash (LSH) indexes such as [69]. Moreover, CMDL also maintains elastic search indexes, e.g., the BM25 index [58], on both the data content and contextual metadata for documents and tabular columns. As will be explained in the next section, these indexes are used as weak supervision signals to construct the joint representation.

---
[3]Unlike min or max pooling that are biased toward and emphasize few values, mean pooling is a more effective representation for the entire set [43]. This is also consistent with the state-of-the-art techniques we compare with in our experiments [15, 31].

# 4 MULTI-MODAL JOINT REPRESENTATION

There are various methods for establishing relationships between unstructured documents and tabular columns, e.g., similarity in data captions, column header and document titles, overlapping content values, and similarity in an embedding space. The sketches presented in Section 3 capture fragments of these relationships. In this section, we address the question of: *How can we construct a common representation that encodes all of these fragments in a meaningful way?* Such representation has thus the potential to outperform the individual sketches in the discovery tasks.

Our goal is to build a joint-representation model that generates embeddings for text documents and tabular columns such that the embeddings are close to each other for similar pairs, and otherwise far from each other. However, the lack of sufficient labeled dataset, which is common in data lake settings, represents a real challenge for constructing such model. In Section 4.1, we first present a solution for creating the needed labeled dataset, and then in Section 4.2, we present the joint representation model.

## 4.1 Training Dataset Generator

In Figure 3, we present CMDL's weak-supervised framework for generating a labeled training dataset. The framework integrates CMDL's indexes in a novel way for data labeling. At first, let us ignore the optional preprocessing phase (surrounded by a red-dotted line), and focus on the main workflow. The process starts by building a random sample from the two different modalities (documents and columns), and then considers the pairs produced from the Cartesian product of the two samples. For pair (doc $d$, col $c$), we need to generate a label (0 or 1) indicating whether $c$ is related to $d$. For that we leverage the Snorkel weak supervision platform [55] while plugging in our custom labeling functions. Let us briefly overview Snorkel platform, and then describe the labeling functions.

**Overview on Snorkel Platform.** We opt for Snorkel because it is the state-of-the-art platform for generating training labels where no true labels exist [54, 55]. Multiple weak sources of supervision, named *labeling functions* (LFs), each providing possibly a weak or inexact label, are used to label the input data points. Snorkel's main strength lies in the fact that these LFs could be imprecise and inaccurate, nevertheless Snorkel has the ability to combine these *noisy* labels and generate higher quality ones through its generative and discriminator models. The generative probabilistic model is fit to estimate the accuracy of the LFs. The model estimates the accuracy and the correlations of the labeling functions using only their agreements and disagreements, and then it reweights and combines their outputs. A single set of probabilistic training labels is generated by this process. These are then put through a discriminator model which runs supervised classification on the input data features. The discriminator ensures that the model generalizes beyond the labeled data points.

**CMDL's Indexes as Labeling Functions.** In our use case, the input to Snorkel is the (doc $d$, col $c$) pair. The labeling functions (LFs) correspond to the different types of indexes we built on the sketches. The probes on each index are carried out for top-$k$ matches, for a small number $k$ to ensure a high quality labels. We provide four main labeling functions as illustrated in Figure 3. (1) The *semantic-based solo embedding* function takes $d$'s embedding and queries the index to retrieve the top-k matching columns. If $c$ is among them, then the function generates 1 (indicating that they are related), otherwise it generates 0 (indicating that they are not related). (2) The *syntactic-based* function generates a minwise hash signature based on $d$'s content, and then queries the LSHEnsemble index for the top-$k$ matching columns. The same idea applies for the keyword elastic search functions including (3) Function for the actual *content similarity*, and (4) Function for the *metadata similarity*.

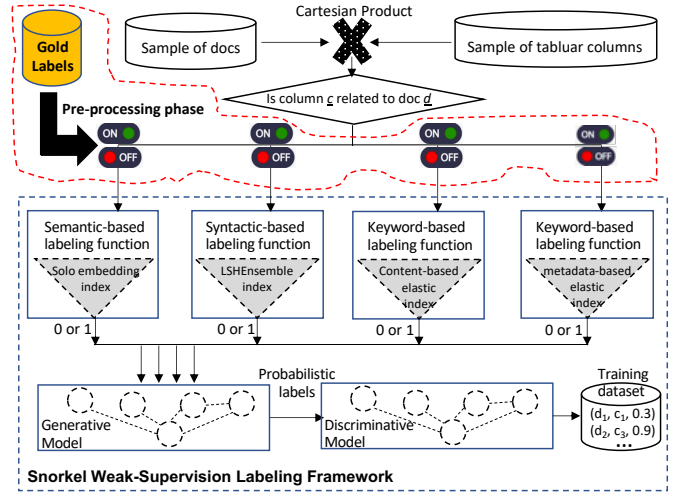

Figure 3: CMDL labeling and training dataset generation (a zoom-in over the green-colored box in Figure 2).

Although, the top-$k$ index probes from the labeling functions may not return all column matches for a given document, this is generally acceptable for deep neural network (DNN) training because: (1) the framework is bounded by the selected samples and thus completeness is not guaranteed any way, (2) DNN training, in the first place, does not require completeness, but instead its power lies in its ability to learn patterns based on a small sample size, and (3) The top-$k$ selection enables the joint model to learn based on the *best matches*, which is intuitively a better choice compared to, for example, random selection from the positive cases. It is also worth mentioning that the indexes associate a similarity score with each returned column object. Therefore, documents that do not have any (or less than $k$) true matches can be detected, and thus the low-quality matches (i.e., below a certain threshold) are eliminated from the generated training dataset.

The binary outputs from the labeling functions are then passed to the Snorkel generative model to generate probabilistic labels for the input pairs. Finally, the discriminator model is trained using the standard cross-entropy loss function over all the document-column pairs from the samples to generate the labeled training dataset with a relatedness degree (between 0.0 and 1.0) for each pair.

The training dataset generation is theoretically quadratic in terms of space and time complexities w.r.t the sampled DEs from the two modalities. However, this process remains practical for the following reasons: (1) It is on a sample, and our experiments show that a small percentage, e.g. 10% sample size, is sufficient for a given data lake, (2) The label generation is a one-time activity and is not part of the online discovery cycle, (3) A single index probe using a document $d$ as input actually labels all columns w.r.t $d$, i.e., all returned columns are labeled with 1 and all other columns are labeled with 0. (4) The generative model only considers those pairs that are labeled 1 by at least one LF; (5) The output labels, as an effect of the above, constitute a sparse matrix and can be efficiently stored in a compressed format.

**Augmented Preprocessing Phase Based on Gold Labels.** The Snorkel framework is expected to eliminate (or reduce) the influence of poor labeling functions compared to the high-quality ones. This tuning takes place automatically especially when tens of labeling functions are provided [55]. However, in applications where the number of labeling functions are limited, which is our case, we observed that poor labeling function(s) remain to have some influence on the system and the accuracy of the final model might suffer.

To remedy this problem, we augment an optional preprocessing phase, subject to the availability of what we refer to as *gold labels* (see Figure 3). Gold labels is basically a tiny labeled dataset (ground truth) that in itself is not enough to guide or generate a supervised model, but can be leveraged to boost the accuracy of the weakly-supervised model. In this phase, we use the gold labels to learn about the accuracy of the different labeling functions. And then we deploy a heuristic that switches off the labeling functions whose accuracy is observed to be below a certain threshold, say 50%, relative to the accuracy of the best labeling function. In concept, this preprocessing phase is generic and applies to all applications facing the aforementioned issue. This approach literally integrates the gold labels into the Snorkel framework, which is different from the standard approach of simply combining (union) Snorkel's output with any available labeled dataset [54]. Our evaluation presented in Section 6 shows that such preprocessing phase makes a noticeable difference on the final model's accuracy in some cases.

In summary, the proposed CMDL's labeling framework inherits the following characteristics: **(1) Extensibility.** It is straightforward to extend CMDL with additional labeling functions, and the entire system would work seamlessly. For example, with the surge of large language models (LLMs), one could think of adding LLM-based labeling functions that take an input pair (doc $d$, col $c$) and return the Boolean output flag based on their relatedness degree. **(2) Scalability-Robustness Balance.** The training framework is based on samples from the underlying domains combined with efficient top-k index probes for the labeling functions, and thus it scales well. Nevertheless, as shown in our experiments over diverse datasets, CMDL is always able to generate labeled datasets that yield a robust joint model that consistently outperforms existing techniques. **(3) Tunability.** The ability to tune the labeling functions by switching them ON or OFF depending on the dataset characteristics captured in a gold-labeled subset.

## 4.2 Joint Representation Learning

In this section, we present the *joint representation learning* module. The goal is to build a model, referred to as the *joint representation model*, that learns to transform the document and tabular column DEs from an initial representation, i.e., the solo embeddings, to new embeddings in a shared joint space. The detailed workflow is depicted in Figure 4.

**Training Loss Function as a Building Block.** Before describing the workflow, let us first describe the training loss function as it impacts the design of the entire workflow. Typically, deep learning models, e.g., neural network models, use either pairwise or triplet loss functions to converge to a configuration where pairs of input objects, e.g., images, with the same labels are closer than those with different labels [60]. However, pairwise loss functions mandate all positive (or negative) pairs to be within a pre-configured positive (or negative) distance range. This is found to highly restrictive and prevents any distortions in the embedding space [46]. Triplet loss functions, on the other hand, are more powerful because they do not use a fixed threshold to distinguish between similar and dissimilar objects. Instead, they can distort the embedding space to accommodate outliers and higher intra-class variance for distinct classes [23]. Therefore, we opt for the triplet loss functions.

The triplet loss function works on a trio of objects at a time; an *anchor* data point (say $x_t$), a *positive* data point that is a good match with the anchor (say $x_{cp}$), and a *negative* data point that is not a match with the anchor $x_{cn}$. Triplet loss will guide the model to re-position the objects in the new embedding space by pushing the negative sample away from the anchor while pulling the positive sample closer to the anchor. that is, the distance from $x_t$ to $x_{cp}$ plus a margin $\beta$ be smaller than the distance from $x_t$ to $x_{cn}$.

$$\mathcal{L}(x_t) = \max\bigl(0, \beta + d(x_t, x_{cp}) - d(x_t, x_{cn})\bigr) \quad (1)$$

**Joint Representation Workflow (Figure 4).** Starting with the labelled training dataset $D$, the *Mini-Batch Generator* acts as a partitioner that generates non-overlapping partitions of DEs, each is called a *mini batch*–the union of these mini batches covers $D$. Each mini batch consists of a small number of randomly selected document DEs (say $m$) and column DEs (say $n$). The ratio between $m$ and $n$ is the same ratio between the total number of documents and columns in $D$.

The next step is to create from a given mini batch $B$ a set of triplets that act as inputs for training the triplet loss model, which is the role of the *Triplet Generator* module. A triplet has the general form of a document $d$ as the anchor point, and two tabular columns $c_i$ and $c_j$ as the positive and negative samples to the anchor. The challenge here is that typically in triplet loss training, an anchor point participates in at most a single triplet within a full epoch (which covers the entire training dataset) [23]. However, in our case, within a mini batch, large number of triplets could be generated for the same anchor, which has the upper bound of $(n/2)^2$ if we consider all combinations of positive and negative column samples. This does not only create potential biases, but also substantially increase the training time. We address this challenge through the procedure depicted in Figure 5.

Basically, a mini batch can be viewed as a small $m \times n$ matrix, where the score within each cell $(d_i, c_j)$ represents the relatedness degree between this pair. Then, based on a pre-defined threshold, we categorize the scores into negative and positive relationships– which is highlighted by the red and green colors in Figure 5, respectively. For each document (i.e., a row in the matrix), the relationships to the $n$ columns (positive and negative) would look like the middle part of Figure 5. Notice that the current encoding of the DEs, which is generated the profiler, is oblivious to the relatedness scores, and thus the positive and negative cases are intermixed in the current representation space.

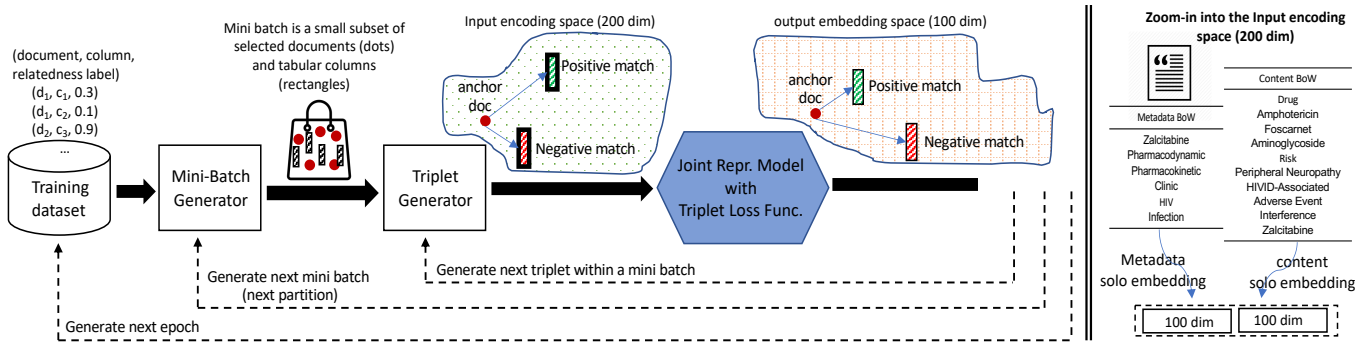

**Figure 4: CMDL workflow for constructing the *Joint Representation Learning* model (a zoom-in over the orange-colored box in Figure 2).**

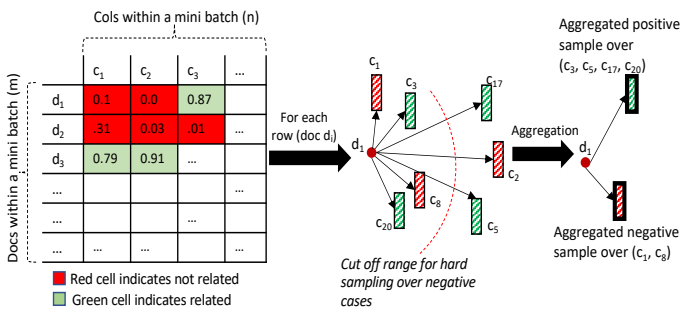

**Figure 5: Example of triplet generation.**

Now, to avoid generating all quadratic combinations of positive and negative triplets, we devise a hard sampling technique [46] augmented with an aggregation step. The hard sampling aims at selecting the hard cases on which we want the model to re-adjust and improve, as such, noisy or weak signals coming from borderline triplets are eliminated. Our experiments show that this approach yields around 10x faster convergence and higher accuracy (Section 6.4). For our scenario, we aggregate all positive samples into one instance (e.g., aggregating columns $c_3$, $c_5$, $c_{17}$, $c_{20}$ as in Figure 5), while selectively aggregating the negative samples within a certain range from the anchor point–those are the hard samples we want to push away (e.g., aggregating only columns $c_1$ and $c_8$ while ignoring $c_2$). This range can be defined based on several criteria, e.g., the $k^{th}$ distance to a positive sample or the average or median over the negative samples. With this aggregation procedure, each document in a mini batch generates only one triplet for the model training [4].

Back to the main workflow in Figure 4, the generated triplets are now fed to the joint representation model along with its triplet loss function to learn a new output representation that better represents the relatedness relationships. The training performed on all mini batches (which covers the entire training dataset) is considered a single epoch. Once finished, another epoch with full random generation of mini batches is executed until the model converges, i.e., the loss across two consecutive epochs is within a small threshold.

---
[4] Documents that do not have both positive and negative sample columns are simply ignored.

The *Joint Representation Model* itself is a deep multi-layer NN with 200 dimensions as input and 100 dimensions as output. For the input side, each DE, either a document or column, has 200 dimensional encoding as a concatenation of two sub-encodings, each in a 100 dimensional space. These sub-encodings represent the solo embeddings for metadata and content generated by the CMDL's profiler (see the right side of Figure 4 for an example of document encoding).

It is worth mentioning that although the input encoding representation takes only the metadata and content into account, the positive and negative relationships that also go into the model (and inferred from the module presented in Figure 3) take additional factors into account including syntactic and keyword-search relationships.

The final step after the model is being trained is to apply it over the all document and column DEs to generate their new representations in the new joint space. The output is fed back to the CMDL's indexing framework to be indexed as depicted in Figure 2.

## 5 KNOWLEDGE GRAPH AND DISCOVERY INTERFACE

### 5.1 EKG and Relationship Types

Similar to the work proposed in [31], all discovered relationships among the DEs pairs and their indexes represent a materialization of the CMDL's Enterprise Knowledge Graph (EKG). The nodes of the graph include tabular columns, tables, and documents (the latter is a new type of DEs not supported by [31]). The edges represent different relationship types, and the strength of a relationship is the edge weight. Relationships that are common between column-column and document-column pairs include syntactic measures such as Jaccard similarity and Jaccard set containment, and semantic measures such as solo embeddings. Relationships that are only specific to column-column pairs include column name similarity and numeric statistics similarity, e.g., min-max range similarity, inclusion, and overlapping values. All of the sketches needed to measure these similarities are built by the CMDL's profiler (in Section 3). Other relationships that are only specific to document-column pairs are the cross-modality joint embeddings presented in Section 4.

Between table-table pairs, higher-order relationships include PK-FK (for joinability) and unionability, which are similar to those proposed in [15, 31, 49]. The PK-FK relationship represents a special type of inclusion dependency between a pair of participating tables (primary table $P$ and foreign table $F$) [24]. The relationship requires the values of a column from $F$ to be entirely contained in a column from $P$. Moreover, the two columns should have similar names, and $P$'s column is estimated as a primary key (i.e., having a cardinality close to 1). Unlike the Aurum system [31] that uses Jaccard similarity as an inclusion measure, CMDL uses Jaccard set containment since it is better aligned with the aforementioned definition and more robust under different cardinalities for PK and FK columns [69].

For the unionability relationship, two tables are considered unionable if the columns from both tables show significant match in name, value containment, numeric range, and semantics. For a given table $T$, we first consider each column in $T$ (say $T.c$) and leverage CMDL's similarity sketches to discover the top-k most unionable columns to $T.c$ based on a combined score (ensemble) over the four measures mentioned above. By performing this step on all columns of $T$, we get a set of candidate tables $\{R_1, R_2, ..., R_m\}$ to investigate. We then apply a maximal graph matching algorithm between the columns of $T$ and the columns of each $R_i$ to compute an overall score for $T$ and $R_i$ unionability. The details of this algorithm is presented in TUS [49], and thus omitted from this paper.

## 5.2 Discovery Interface

CMDL adopts the same SRQL engine proposed in Aurum [31] with several extensions specific to our system. SRQL engine has two concepts: Discoverable element (DE) and Discovery primitive (DP). While the DEs correspond to the nodes of EKG, DPs correspond to the APIs over the DEs. The SRQL supports three types of queries including *catalog-based* (searching based on DEs metadata), *content-based* (searching based on DEs content), and *relationship-based* (searching and navigating the graph based on the relationships among the DEs). The results returned from these queries are in the form of *Discovery Result Sets* (DRS) associated with APIs for manipulation and provenance tracking.

For the purpose of CMDL, we extend SRQL by adding the new DEs of document type, and new DPs for searching within the document modality as well as others for searching across the document and column modalities. Further, we add different compositions of the DRS obtained from the DPs that include normalized sum of similarity scores from different relations between a pair of DEs.

**Example.** *Referring back to our motivation scenario from Section 1, the discovery questions highlighted in Figure 1 can be expressed as (the syntax is mostly self-explanatory):*

Q1: Retrieve documents related to enzyme "Thymidylate synthase":
$r1 =$ **content_search**(value: "Thymidylate synthase", mode: Text)
*The second argument specifies the scope of search to be the text documents. $r1$ is now a DRS consisting of array of documents.*

Q2: Find tables related to the $1^{st}$ returned document:
$r2 =$ **crossModal_search**(value: r1.[1], topn: 3)
*Alternatively, instead of searching using the entire document, the end-user can only supply the yellow-highlighted text (as in Figure 1) inside the "value" field. crossModal_search() is a new API specific to CMDL. $r2$ is now a DRS consisting of array of table names.*

Q3: Find tables related to the $3^{rd}$ returned document:
$r3 =$ **crossModal_search**(value: r1.[3], topn: 3)
*This is the same as Q2, and $r3$ now consists of [Enzyme_Targets, Drugs] table names.*

Q4: Find tables joinable with "Drugs" table:
$r4 =$ **pkfk**(value: r3.[2], topn: 2)
*This API discovers the two top ranked joinable tables with "Drugs". $r4$ now consists of [Enzyme_Targets, Drug_Interactions].*

Q5: Find tables unionable with "Enzyme_Targets" table:
$r5 =$ **Unionable**(value: r4.[1], topn: 2)

## 6 EVALUATION

In this section, we evaluate the CMDL system for the three key discovery tasks listed in Section 2, namely $\text{Doc}_{to}\text{Table}$, $\text{Table}_J\text{Table}$, and $\text{Table}_U\text{Table}$. The experiments are carried out on a 32 core machine with 32GB RAM and 4TB SSD storage. As will be explained in this section, we leverage several existing benchmarks from literature in addition to adding new benchmarks, especially for the cross-modality discovery.

**Baselines.** For the $\text{Doc}_{to}\text{Table}$ discovery, we characterize the prior approaches as either keyword search or similarity sketch techniques [15, 31, 69]. While the former extracts keywords from documents to query elastic search indexes built on the tabular columns, the latter constructs succinct signatures, e.g., minwise hashing sketches, for indexing and comparison. In the experiments, we use the BM25 engine for elastic search [58], and LSHEnsemble hash index for containment similarity [69]. For the structured data discovery tasks including $\text{Table}_J\text{Table}$ and $\text{Table}_U\text{Table}$, we compare with two of the state-of-the-art discovery systems, more specifically Aurum [31] and D3L [15]. Aurum materializes the schema and content similarity links between column pairs in the form of a knowledge graph, which can then serve different discovery queries. D3L builds hash-based signature sketches on multiple fine-grained signals for column similarity, which are combined at query time to find the top-$k$ most related columns or tables.

**Test Suite.** We use three data lakes in the evaluation (refer to Table 1). The first, called Pharma, consists of a tabular data from pharmaceutical databases (DrugBank, ChEMBL, and ChEBI) in addition to a text corpus of PubMed and MedLine abstracts. These abstracts correspond to the citations provided from within the DrugBank database, e.g., each row in tables like Drug and Enzyme contains a reference to a related abstracts. The second is a collection of government open data collected in CSV format. This lake corresponds to the *Smaller Real* testbed used in D3L [15] and is referred to as UK-Open in this paper. The third, called ML-Open, contains a set of ML datasets from open data portals such as Kaggle and OpenML, which is reported in [35]. This data lake has three variations depending on the sizes of the files: Small Scale (SS), Medium Scale (MS), and Large Scale (LS). Table 1 summarizes the three data lakes. Each data lake consists of both structured and unstructured data collections. Both the CSV and MySQL formats indicate tabular data, while Text indicates unstructured text documents. The table lists important statistics for each data lake such as the number of tables, columns, and documents as well as the range of file sizes.

Table 1: Overview of the evaluation datasets.

| Data lake | Data collection | Format | Num. of tables | Num. of DEs[∓] | File sizes | Numeric attributes |
|---|---|---|---|---|---|---|
| Pharma | DrugBank | CSV | 82 | 418 | 0-400MB | 7% |
| | ChEMBL | MySQL | 77 | 543 | 0-300MB | 41% |
| | ChEBI | MySQL | 10 | 61 | 0-500MB | 34% |
| | PubMed | Text | - | 2000 | 0-4kB | - |
| | DrugBank-Synthetic | CSV | 80 | 220 | 0-10MB | 7% |
| UK-Open [15] | Govt. data | CSV | 654 | 8766 | 0-200MB | 18% |
| | Synthetic text | Text | - | 2360 | 0-2kB | - |
| ML-Open [35] | Small Scale (SS) | CSV | 28 | 243 | 0-1MB | 33% |
| | Medium Scale (MS) | CSV | 159 | 1286 | 0-10MB | 46% |
| | Large Scale (LS) | CSV | 46 | 2550 | 0-100MB | 69% |
| | Reviews | Text | - | 1500 | 0-7kB | - |

[∓] The number of DEs indicates the number of columns for tabular datasets, and the number of documents for Text datasets.

Table 2: Overview of the evaluation benchmarks.

| Discovery Task Category | Benchmark | Data Lake | Data Sets | #Queries | Average Answer Size | mQCR[★] | Ground Truth Generation |
|---|---|---|---|---|---|---|---|
| $Doc_{to}Table$ [‡] | 1A | UK-Open | Synthetic text + Govt. data | 2360 | 55 | .05 | Synthetic |
| | 1B | Pharma | PubMed + DrugBank | 927 | 8 | .006 | From the database |
| | 1C | ML-Open | Reviews + MS | 1500 | 7 | .003 | Manual |
| $Table_J Table$ (syntactic join) | 2A | UK-Open | Govt. data | 1000 | 17 | .62 | [15] |
| | 2B | Pharma | DrugBank [‡] | 147 | 8 | .08 | Brute force |
| | 2C | ML-Open | SS | 150 | 6 | .71 | Brute force |
| | | | MS | 690 | 6 | .45 | Brute force |
| | | | LS | 790 | 6 | .02 | Brute force |
| $Table_J Table$ (PK-FK join) | 2D | Pharma | DrugBank [‡] | 1 | 55 | .28 | Manual |
| | | | ChEMBL | 1 | 96 | .25 | From schema def. |
| | | | ChEBI | 1 | 9 | .22 | From schema def. |
| $Table_U Table$ | 3A | UK-Open | Govt. data | 654 | 110 | .5 | [15] |
| | 3B | Pharma | DrugBank-Synthetic [‡] | 80 | 15 | .23 | Synthetic |

[★] mQCR means Median Query Cardinality Ratio and is explained in the main text.

[‡] Indicates the benchmarks newly contributed with this work. They are made publicly available for the research community.

On top of these data lakes, we use the nine benchmarks listed in Table 2 for evaluation. As listed in the table, the benchmarks are divided to serve different discovery tasks. In the table, we list the data lake corresponding to each benchmark, and the collection of datasets used in the evaluation. For the $Doc_{to}Table$ benchmarks, i.e., {1A, 1B, 1C}, the query consists of a document, and the discovery task is to return the top-k related tables, for a given $k$. All evaluated methods under this category first compute the relatedness scores based on the individual tabular columns, and then aggregate these scores to the table level. The number of queries (i.e., the $5^{th}$ column in Table 2) corresponds to the number of documents in the referenced data lake that appear in the ground truth, i.e., documents with at least one link to a tabular column.

The benchmarks related to the joinability discovery task are subdivided into *syntactic* and *PK-FK join* types as in [31]. The former type discovers the potential joinability over any pair of columns from two distinct tables, while the latter limits the discovery to only the foreign keys related to a given primary key. For the syntactic join benchmarks (2A, 2B, 2C), one query is submitted for each column that appears in the ground truth (Say $T.c$), and the discovery task is to find potential joinable columns in tables other than $T$. For the PK-FK discovery benchmark (2D), the value in the *Average Answer Size* column represents the total number of the PK-FK links in the corresponding dataset (i.e., the ground truth). We use a single query for discovering these links as performed in [31]. Finally, for the benchmarks related to the $Table_U Table$ discovery task, i.e., {3A, 3B}, the goal is to discover for a given table $T$ other tables potentially unionable with $T$. The number of queries equals the number of tables in the referenced data sets.

As highlighted in the table, the ground truth for the benchmarks are obtained using different methods. Benchmarks that consume synthetic datasets (i.e., 1A and 3B) have their ground truth generated based on the algorithms generating the datasets. Several ground truth datasets are generated by simply running brute force algorithms on the search space (e.g., for Benchmarks 2B and 2C), while others are obtained from the datasets themselves (e.g., for Benchmarks 1B, 2D-ChEMBL, and 2D-ChEBI). The ground truth for the UK-Open benchmarks (2A and 3A) are obtained from literature, while the rest are manually annotated.

The last two columns in Table 2, i.e., the *Avg answer size* and *median query cardinality ratio (mQCR $\in [0, 1]$)*, are computed based on the ground truth datasets. The query cardinality ratio is computed as follows. Assume Benchmark 1A in which a query document $q$ is transformed into a bag of words (say consisting of 7 words). In the ground truth dataset $q$ is related to a tabular column $c$ (say containing 100 values), then the QCR for the link $q$-$c$ is 7/100 = 0.07. The mQCR presented in Table 2 is calculated as the median over all links in the ground truth. The mQCR reflects the skewness degree between the cardinality of the query DE and the discovered DEs. As we will show later, CMDL is more robust and outperforms other systems under high skewness (i.e., small mQCR).

**Evaluation Metrics.** The accuracy of the discovery operations is measured by the standard *precision* and *recall* metrics computed based on a top-$k$ matches for a given query. The system's profiling overheads are measured by the wall clock time and storage size, while the model training till convergence is measured by the number of epochs, wall clock time, and error percentage.

**Default Settings.** Unless otherwise explicitly stated, the following default parameters' settings are used throughout the different experiments. The sample size for CMDL's labeling and training dataset generation (Figure 3) is set to 10%. The size of the gold labels (Figure 3), when applicable, is set to 10% of the ground truth size. The mini-batch matrix size $m \times n$ (Figure 5) is set to 8%, i.e., the number of documents $m$ and columns $n$ in a single mini-batch is 8% of the number of the corresponding DEs in the training dataset. The hard sampling strategy is enabled by default, and its cutoff threshold is set to the average distance over all negative samples (Figure 5). Finally, the triplet loss margin $\beta$ (Eq. 1) is set to 0.2. The impact of varying these parameters and changing their default settings is studied in Section 6.4.

## 6.1 Document-to-Table Discovery

We use Benchmarks 1A, 1B, and 1C for these set of experiments as illustrated in Figure 6. We evaluate three variations of CMDL (prefixed with "CMDL") including: (1) leveraging only the solo embeddings created from the profiler, (2) leveraging the new joint embeddings that brings the two modalities into a joint vector space, and (3) augmenting #2 with the fine-tuned model for labeled data generation (i.e., the tuning based on gold labels as presented in Section 4.1). We compare with three families of baselines including: (i) containment-based sketches that leverage minwise hashing, (ii) elastic search algorithms under four settings (the top four labels in Figure 6), namely BM25 (TF/IDF), which is the default similarity measure, and LM Dirichlet measure over the union of content values and schema information (the $1^{st}$ and $2^{nd}$ labels), and BM25 over the content values, and the schema information separately (the $3^{rd}$ and $4^{th}$ labels). And (iii) entity matching algorithms (the right-most two labels in Figure 6) using the standard SpaCy model [9] using two similarity metrics; Jaccard and Jaro. There is a customized model for SpaCy, called SciSpaCy [10, 50], that is fine-tuned on the same dataset we use in Benchmark 1B (PubMed), and hence we use SciSpacy for the 1B experiment.

The queries in the benchmarks are top-k queries, i.e., the precision and recall are computed relative to the returned top $k$. Therefore, we repeat each benchmark and each method for different $k$ values. For each benchmark, the range of $k$ varies to cover a spectrum around the average answer size for that benchmark (refer to the $6^{th}$ column in Table 2). The caption of Figure 6 lists the ranges of $k$ used for the different benchmarks. With respect to varying $k$, we observed the expected behavior, i.e., as $k$ increases the recall goes up and the precision goes down. This is true across the three benchmarks. That is why we eliminated $k$ as a separate dimension for better visualization, and only highlighted few $k$ values inside the figures to show the trend.

The key observations from Figures 6(a), (b), and (c) can be summarized as follows.

*Elastic Search Performance.* Elastic search-based techniques are not reliable and their accuracy is highly dependent on the characteristics of the benchmark data. For example, schema-based search does not produce any promising results across the board. The reason is that the metadata alone is very short, and thus this method misses the right relationships. The other three techniques perform relatively well for Benchmark 1B, where drug names are very unique and can easily connect objects together, but they fail to maintain competitive performance for the other benchmarks. Especially the recall for these techniques is always very low even under larger $k$. This is because these methods do not capture any semantics, and many of the links they discover are not real relationships in the ground truth.

*Entity Matching Performance.* The entity extraction and matching techniques treat each tuple in the tabular data as a document. They, then discover a relationship between an unstructured document $d$ and a table $T$ if they established an entity-matching connection between $d$ and any tuple in $T$. From the results in Figure 6, we observe that unless these techniques are highly fine-tuned to discover and extract meaningful entities from a specific domain, their extraction quality becomes very poor, which results in near-random relationships (e.g., Benchmarks 1A and 1C). For Benchmark 1B, the fine-tuned SciSpaCy model improves the performance to a competitive level. Nevertheless, CMDL's variations still outperform by a noticeable margin. Notice that for Benchmark 1B, the Jaro-based algorithm was not feasible to compute due to the quadratic time complexity involved in the Jaro function (the estimated finish time was 10+ days).

*Containment Search Performance.* The containment-based search is shown to perform better than some of the elastic search and entity matching algorithms under the three benchmarks. However, its performance is not predictable and it is consistently far below CMDL's performance. One of the reasons is that LSHEnsemble index is threshold based, and therefore it is incapable of producing meaningful ranked results. This is the reason behind the unexpected reverse trend w.r.t the precision measure under Benchmark 1A.

*CMDL's Performance.* Compared to the baselines, we observe that CMDL's variations show more promising results and better tradeoffs between precision and recall across the different benchmarks. CMDL's variations are consistently superior, which indicates that it is domain agnostic, and that its sub-models (i.e., the label generation or joint representation models) are able to learn relationship patterns effectively. Clearly, for Benchmark 1B, which is the easiest benchmark due to the uniqueness of the drug names, the different techniques become closer to each other, but still CMDL's accuracy remains the skyline to the rest.

Comparing CMDL's variations, we observe that the new joint representations perform around 5% to 10% better over the solo embeddings–thanks to the integration of multiple signals into one representation. Notice that Figure 6(c) shows the phenomena mentioned in Section 4.1 where highly imprecise labeling functions could cause harm to the joint representation model. Nevertheless, the CMDL's variation with gold label tuning brings the joint representation model back as the best method (more analysis on this tuning is presented in Section 6.4). Such 5% to 10% improvement may seem not significant at first sight, however, given the problem complexity as evident by the poor and/or inconsistent behavior of the other techniques, it is one step forward. More importantly, it represents promising results due to the big functional edge of the joint representation model over solo embeddings, i.e., the former is an extensible framework within which new and additional similarity metrics such as LLM-based functions can be seamlessly integrated (as discussed in Section 4.1), whereas the latter is a terminal feature and not subject to such extensions.

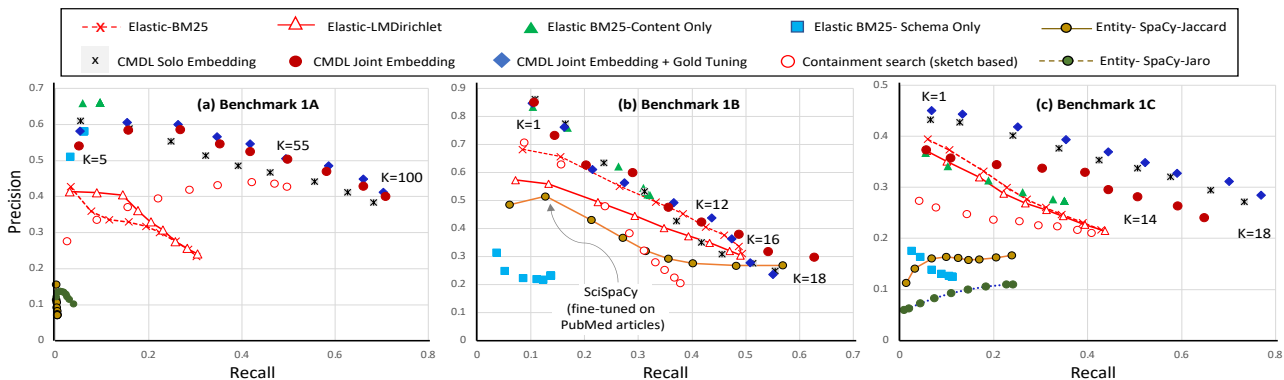

Figure 6: Effectiveness of the cross-modality data discovery. The range of $k$ for the top-k queries differs based on the benchmark. For Benchmark 1A, $k \in \{5, 15, 25, 35, 45, 55, 70, 85, 100\}$, and for Benchmarks 1B and 1C, $k \in \{1, 2, 4, 6, 8, 10, 12, 14, 16, 18\}$.

## 6.2 Joinable Tables Discovery

Similar to studies in other systems [15, 31], we consider two types of join discoveries, namely *syntactic* and *PK-FK*. Table 3 compares CMDL with Aurum and D3L on the syntactic join discovery. In this experiment, we set $k$ (the size of the discovered answer set for each query) to match the ground truth size. As a result and as highlighted in Table 3, the precision and recall scores become identical.[5] The syntactic join discovery is tested using Benchmarks 2A, 2B, and 2C. While the ground truth for Benchmark 2A is manually annotated, the ground truth for Benchmarks 2B and 2C is generated by running an expensive all-pairs exact set similarity.

The key difference between CMDL and the other systems is that CMDL uses the Jaccard set containment measure compared to the Jaccard similarity measure used by the other systems. It is clear that CMDL outperforms the other systems in most cases. That is mainly because the set containment is more robust under skewed cardinalities of the joined DEs. Recall that the *mQCR* ratio (the last column in Table 2) indicates such skewness degree, e.g., Benchmarks 2B and 2C-LS are highly skewed as indicated by their low *mQCR* score. Interestingly, Benchmark 2A exhibits poor accuracy on all systems despite its lower skewness. This is because the ground truth is generated by manual annotation and does not necessarily imply high syntactic overlap.

For the PK-FK join discovery, CMDL computes potential join relationships using a combination of data containment, schema overlap,

---

[5]This measure is referred to as 'R-Precision' in [25].

Table 3: Evaluation of syntactic join discovery.

| Benchmark | Workload | Precision = Recall | | |
|---|---|---|---|---|
| | | Aurum | D3L | CMDL |
| 2A | Govt. data | .22 | .22 | .30 |
| 2B | DrugBank | .21 | .37 | .62 |
| 2C | SS | .70 | .70 | .70 |
| | MS | .55 | .55 | .57 |
| | LS | .21 | .21 | .44 |

Table 4: Evaluation of PK-FK join discovery (Benchmark 2D).

| Database | Known PKFKs | Aurum precision/ recall | CMDL precision/ recall |
|---|---|---|---|
| DrugBank | 55 | .58/.36 | .33/.91 |
| ChEMBL | 96 | .09/.53 | .24/.59 |
| ChEBI | 9 | .71/.58 | .71/.58 |

and data cardinality profiles. The ground truth in Benchmark 2D is based on the schema definition for the databases ChEMBL and ChEBI. However, no foreign key constraints are defined for the DrugBank tables, and thus we created its ground truth manually.

Table 4 compares CMDL with Aurum–D3L does not compute PK-FK links. On DrugBank, CMDL exhibits a high recall due to the use of the Jaccard set containment instead of the Jaccard similarity used by Aurum. Nevertheless, CMDL has less precision (more false matches) because the true cardinality of the many identified *key* attributes is slightly lower than 1. This is caused by duplicates in the database stemming from a lack of enforcement of key constraints. Both Aurum and CMDL exhibit a low recall on ChEMBL which, as suggested in Aurum [31] is due to many semantically reasonable joins that are not included in the schema. CMDL achieves a better accuracy because of its use of the schema similarity filters. For the ChEBI database, all PK-FK constraints are defined on numeric columns. Since CMDL uses the same numeric overlap similarity measure used in Aurum, the results from both systems are identical.

## 6.3 Unionable Tables Discovery

We use two workloads in this evaluation: Benchmark 3A, which is based on the UK-Open data lake provided by the D3L system [15],

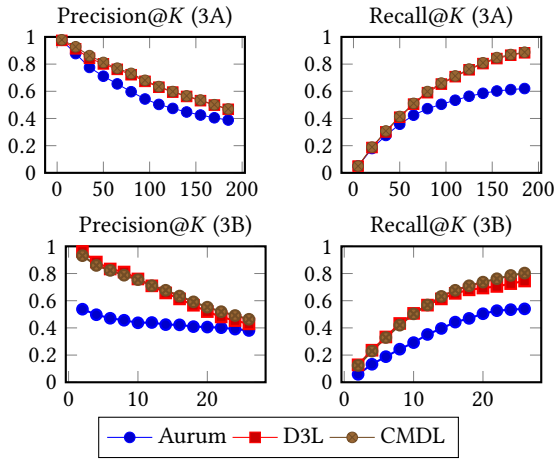

Figure 7: Evaluation of unionable table discovery.

Table 5: Comparing individual similarity metrics

| Bench-mark | Metric | name | containment | numeric | semantic | CMDL ensemble |
|---|---|---|---|---|---|---|
| 3A | RR | 0.82 | 0.63 | 0.34 | 0.62 | 0.83 |
| | Queries answered | 99% | 99% | 87% | 100% | 100% |
| 3B | RR | 0.44 | 0.65 | 0.04 | 0.73 | 0.79 |
| | Queries answered | 75% | 100% | 20% | 100% | 100% |

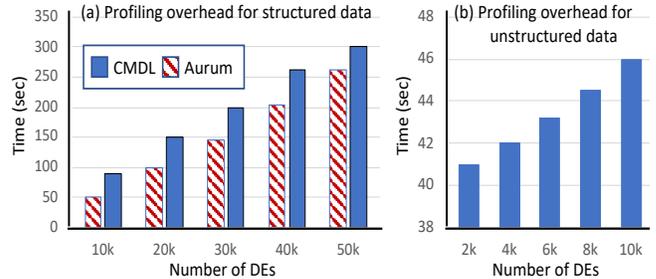

Figure 8: CMDL profiler overheads.

and Benchmark 3B, which is synthetically generated from the DrugBank tables. The synthetic tables are generated by performing multiple projections and selections on the base tables–a process similar to that used in [49]. The results are presented in Figure 7.

Both the CMDL and D3L systems combine multiple scores to detect unionability, but in different ways. For a given pair of columns, both systems calculate scores for the similarities of column names, set containment, numeric overlap, and semantic matching. CMDL uses an ensemble measure wherein the scores of individual similarity measures are combined first before aligning tables using bipartite graph matching. D3L employs a different mechanism in which it first obtains the most unionable tables based on an individual measure before combining the similarity scores (distances) with a weighted euclidean sum. Even though the process of the two systems are different, both essentially combine different evidences for table unionability in a manner that gives more weight to the most discriminating evidence. The results show that both systems perform equally well for the unionability task. Nevertheless, CMDL's superior performance in the PK-FK discovery (Table 3) as well as its unmatched capability for the cross-model discovery (Figure 6) represent value-added benefits over the D3L system. The Aurum system, on the other hand, materializes the unionability relationships based on only two similarity measures, namely *schema similarity* and Jaccard similarity, and combines them by taking the maximum of the two scores. As illustrated in Figure 7, this strategy does not perform as well compared to the other two systems.

**Analysis of individual similarity measures.** We analyze the impact of the individual similarity measures on the unionable table discovery in CMDL using the *Relative Recall* (*RR*) metric [59]. The *RR* score for a similarity measure *S* is the ratio of the size of true matches identified by *S* to the size of the true matches identified by the union of all measures. Table 5 lists the scores along with the number of queries which found at least one true match with the similarity measure. It can be observed that the different benchmarks utilize the four measures differently. For example, for benchmark 3A, the highest scores come from **name** followed by **semantic**, whereas for benchmark 3B, the highest scores come from **semantic** followed by **containment**. This result emphasizes the robustness of the CMDL's **ensemble** measure to the workloads with varying data characteristics (the last column).

## 6.4 System efficiency analysis

**Data Profiling Performance.** Starting with the profiling efficiency for structured data, we compare CMDL's data profiler with Aurum. Since the source code of D3L is not publicly available, we only refer to its performance numbers from [15]. For the purpose of this experiment, we use the UK-Open data lake. In its original form, the data collection contains 654 tables and 8766 columns (as listed in Table 1). To stress test on scalability, we create multiple copies of the data to generate five configurations containing the DEs ranging from 10$k$ to 50$k$. These correspond to 2GB-10GB data on disk. Hash-based content sketches were configured with 512 hashes. A 300-dimensional *fasttext* word embedding model is loaded to memory once.

Figure 8(a) depicts the comparison results. CMDL achieves linear scalability by exploiting the available parallelism in profiling the datasets in a manner similar to Aurum. Nevertheless, CMDL consistently requires a delta amount of extra time compared to Aurum. This is because CMDL constructs more types of data sketches (e.g. vectors for the solo embeddings) and maintains finer granular features (e.g., using word tokens instead of instance values). The same observation is reported for the D3L system [15].

For the profiling performance over the unstructured documents, CMDL deploys an NLP pipeline, which is written in the Gensim python library [56], to build a BoW representation before constructing the content sketches. In this experiment, we use the movie summary documents corpus from the ML-Open data lake. It consists of 1,500 documents, which we replicated few times to scale to 10,000 documents. Figure 8(b) shows that the profiler is super fast, and it can process around 10$k$ documents in less than a minute.

**Training Overhead for Joint Representation Model.** In order to train the joint representation model, we first need to create the sketches by the data profiler. The overheads associated with this

Table 6: Query throughput for different labeling functions.

| Labeling function | Index | Throughput (Qps) |
|---|---|---|
| Content search | Elastic search [36] | 75 |
| Containment | LSHEnsemble [69] | 120 |
| Semantic | Annoy [14] | 1000 |

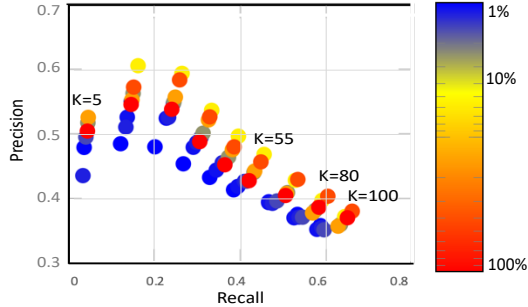

(a) Sampling effect on Benchmark 1A

|  | Gold Dataset % | Schema Search | Content Search | Containment Search | CMDL Solo Embedding |
|---|---|---|---|---|---|
| Benchmark 1A | 1% | 0.36 | 0.38 | 0.31 | **0.42** |
|  | 5% | 0.44 | **0.55** | 0.41 | 0.51 |
|  | 10% | 0.42 | **0.61** | 0.36 | 0.48 |
| Benchmark 1B | 1% | 0.31 | 0.35 | 0.44 | **0.47** |
|  | 5% | *0.17* | 0.51 | 0.38 | **0.52** |
|  | 10% | *0.15* | **0.48** | 0.25 | 0.4 |
| Benchmark 1C | 1% | 0.24 | **0.35** | 0.21 | 0.34 |
|  | 5% | *0.12* | 0.34 | *0.21* | **0.45** |
|  | 10% | *0.14* | 0.29 | *0.2* | **0.41** |

■ Highest Scoring Function   ■ Eliminated Function

(b) Impact of gold labels size on the elimination of weak labeling functions

**Figure 9: Impact of sampling and gold label sizes on the training dataset generation.**

process are reported in Figure 8. Next, the overheads of probing the indexes, as part of leveraging the labeling functions in the weakly supervised framework, are reported in Table 6. We use the throughput metric measured by the number of completed queries per second (Qps). We use a locally hosted BM25 elastic search engine with the out-of-box settings. The nearest neighbor search in semantic metric spaces is lightning fast due to the C++ based memory mapped index used by the tool. Snorkel's label generation model is also very efficient, e.g., the generative model processes 10$k$ data points in one second. Finally, the joint representation model, which is trained using the triplet margin loss, is also implemented in PyTorch. As will be presented later in this section, it converges in around 200 epochs in less than 2 minutes.

**Impact of Sampling and Gold Label Sizes.** As presented in Section 4.1, we create the weakly-supervised training dataset based on a sample, where the default sample size in all experiments is set to 10%. In the experiment presented in Figure 9(a), we use Benchmark 1A to investigate the impact of the sample size on the discovery accuracy. It can be observed that sample sizes in the range of 5%-10% of the input data are sufficient to build a highly effective model. This shows that the generation of the labeled training dataset in CMDL is not an expensive process and scales well. Similar to the experiments presented in Figure 6, the results in Figure 9(a) are obtained under various values for $k \in \{5, 15, 25, 35, 45, 55, 70, 85, 100\}$, where each group (cluster) of points roughly corresponds to a specific $k$ as highlighted in the figure.

Also, the training dataset generation process benefits from gold labels (if available) to improve the labels' accuracy by eliminating imprecise labeling functions (refer to Section 4.1). By default, the gold labels' size used in the evaluation in Figure 6 is 10% of the ground truth dataset. In the experiment presented in Figure 9(b), we vary this percentage to 1%, 5%, and 10%. We observe that while 1% is too small to differentiate among functions, the 5% and 10% sizes were roughly consistent in detecting and eliminating the imprecise functions (highlighted in red) whose scores are less than 50% of the highest score (highlighted in green).

**Impact of Triplet Generation Parameters.** The learning of the cross-modal joint representation relies on triplet generation and triplet loss function (refer to Figures 4 and 5). In the experiments presented in Figure 10, we study in more detail the parameters involved in these processes.

In Figure 10(a), we investigate the impact of the mini-batch size on the model's convergence in terms of both the number of epochs and wall clock time. The default size used in all other experiments is 8%, i.e., the number of documents ($m$) and columns ($n$) in a mini-batch matrix (Figure 5) is 8% from the corresponding DEs in the training dataset. As can be observed from Figure 10(a), a percentage between 5% to 8% is the sweet spot within which the model converges in around 200 epochs in around 2 mins.

In Figure 10(b), we fix the mini-batch size to 8% and the number of epochs to 210, which is the optimal number from the previous experiment, and then vary the hard sampling settings used in the triplet generation algorithm. In one setting (the right-most bar), we disable the hard sampling strategy, and create for each document (anchor point) all possible triplets with each possible pair of a positive and a negative sample. Clearly, this setting results in a huge number of triplets that significantly increases the training time, and worse yet it produces a less accurate model, where the error percentage after the 210 epochs is around 7.34%.

With the hard sampling strategy enabled, we experimented with two methods to calculate the cutoff threshold (refer to Figure 5), namely the average-based and median-based on the negative samples related to a given document. As can be observed, the difference is negligible. Our default setting, which is used in all other experiments, is based on the average computation. Finally, the triplet loss margin ($\beta$) in the triplet loss function (Eq. 1) is tested under different values. As illustrated in Figure 10(c), we observed that when $\beta$ is set to a low value in the range of 0.1-0.3, it produces the best generalization for the model. This is consistent with the results reported in [47].

**End-to-End Usability Study.** In the previous experiments so far, we studied the primitive operations of CMDL from various aspects. These operations can be intermixed to create discovery pipelines of different complexities. In the following experiment, we aim to extract some insights from analyzing an end-to-end discovery pipeline, and for simplicity, we use the 5-step pipeline from our motivation scenario in Figure 1. In this usability experiment, we focus on analyzing the end-to-end execution time and how that time is divided between the system and the analyst (see Figure 11). Recall that the pipeline consists of five tasks: (1) keyword search to retrieve $K$ documents related to a given keyword, (2) $Doc_2Table$ operation to retrieve $K$ tables related to one selected document from #1 output, (3) a second $Doc_2Table$ operation to retrieve $K$ tables related to another selected document from #1 output, (4) $Table_J Table$ to retrieve $K$ tables joinable with one selected table from either #2 or #3 outputs, and (5) $Table_U Table$ to retrieve $K$ tables unionable with one of selected table from #4 output. The numbers reported in Figure 11 are for $K = 3$.[6]

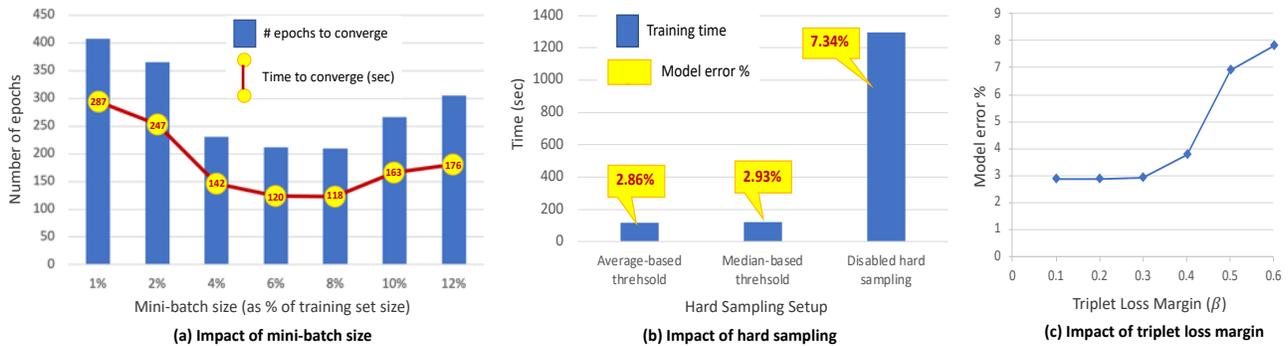

Figure 10: Impact of triplet generation parameters.

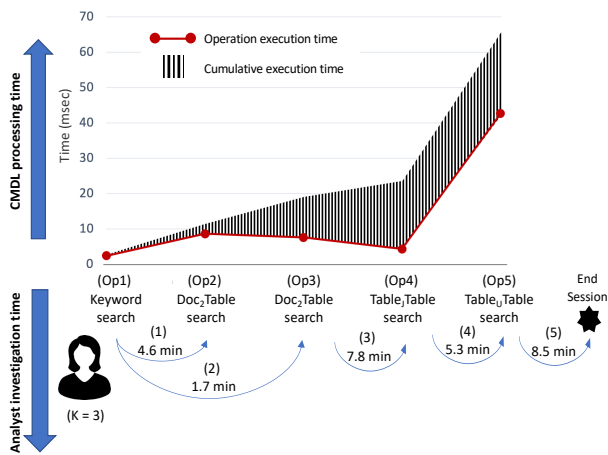

Figure 11: End-to-end study for a discovery pipeline (from Figure 1).

The study involved three domain experts from the biomedicine sciences, and they worked on the Pharma data lake–thus the five discovery tasks in the pipeline are drawn from Benchmarks 1B, 2B, and 3B. Each expert is given two distinct keywords to start with, and the numbers presented in Figure 11 are the average over the six runs. In between the five operations, each expert needs to analyze and process the output from the previous operation, e.g., read the documents from $Op1$ or check the schemas and run few search queries on the tables from $Op3$ and $Op4$, and then select an object of interest for the subsequent operation.

The key observations from the reported results are: (1) CMDL's operations execute in the range of milliseconds and the cumulative system's execution for the entire pipeline is around 65 milliseconds (the upper half of Figure 11(a)). (2) The unionability operation is the most expensive because the EKG is materialized at the column level, whereas unionability requires aggregation to the table level and involves a slightly expensive graph matching algorithm among the columns of each candidate table pairs. (3) Although the system's execution time is clearly minimal compared to the analyst's intervention and investigation time, the former must remain in the range of milliseconds for the system to be interactive and useful to end-users. (4) Analysts indicated that it would be useful if the system can provide a summarization/hint on its output to cut down their investigation time. For example, for the joinability operation ($Op4$), where the system reports the top $K$ joinable tables to a given table, the analyst needs to manually retrieve the schema of each of these tables, examine what new/additional columns or features each table provides over the input table, and possibly check whether or not the tuples containing the search-initiated keyword are actually participating in the join.

## 7 RELATED WORK

**Data Discovery over Structured Datasets.** There has been a lot of recent work attempting to automate discovery tasks over a data lake of relational tables, e.g., schema matches, join dependencies, unionability, etc. These approaches leverage a wide variety of similarity signals such as exact value overlap [67], schema similarity [49], approximate hash sketches [34, 69], ontology matches [33, 57], transformations [11, 68], embeddings [30, 61], statistical profiles [35], with various degrees of combinations among them such as Aurum [31], D3L [15], and TURL [27] systems. However, these techniques and systems do not extend to discoveries across the two modalities of structured and unstructured data.

**Metadata Catalogs and Data Fusion Systems.** One line of work in data discovery is the catalog-based systems that rely on metadata search to discover relationships [39, 44, 52, 65]. For example, Google Goods [38] collects relatively simple metadata about datasets and exposes it through a service. In the last few years, a number of systems including Lyft's Amundsen, LinkedIn's Datahub, Netflix's

---
[6]We also experimented with $K = 5$. The impact on the system's execution time is negligible, while the analyst time for analyzing the outputs has increased to [7, 2.3, 9.6, 7.7, 10.3].

Metacat, Uber's Databook, Airbnb's Dataportal have been proposed to tackle the problem of managing the data lake catalog. These systems primarily operate as metadata hubs where they store and expose metadata about datasets in a data lake. However, these techniques do not support the type of data discovery tasks addressed in this paper.

Data fusion systems, e.g., Google Cloud Fusion [5], provide a generic scalable infrastructure for users to build their own transformation, integration, and discovery pipelines. However, in itself, Google Fusion does not provide the solution to the discovery problem addressed in this paper. In other words, CMDL's modules proposed in Figure 2 still need to be devised, and then the execution can be done on the Google Fusion platform. Another related system is the IBM Watson Discovery, which is an intelligent document processing engine [6]. However, this system is highly tailored towards text documents processing rather than structured data. For example, the system simply treats each tuple in a table as a separate document represented in JSON format. As such, our benchmark datasets, which are close to $16 \times 10^6$ tuples, would require an enterprise-level plan with extremely high cost just to upload the data. Moreover, the system does not support semantic search nor cross-modal discoveries, and thus it cannot substitute CMDL's proposed functionalities.

**Data Search over Unstructured Datasets.** There is no doubt that search techniques over unstructured data have advanced significantly over the years [22, 41, 70]. This is especially true with large language models that are getting closer to mimicking human reasoning about data [18, 20, 29]. However, these techniques are tailored towards specific tasks such as web search, question/answering, conversational dialogues, and content generation [20, 22]. These systems have not been used before for the types of data discovery tasks targeted in this paper. Certainly, our system can leverage these techniques for supporting questions like $Q_1$ in the motivation example (in Figure 1), but beyond that CMDL adds a clear value.

**Data Discovery Across Structured and Unstructured Datasets.** Entity matching is a branch of data discovery that tries to discover and connect entities from structured, semi-structured, and unstructured data, e.g., [13, 21, 53]. For example, ConnectionLens [13] constructs a knowledge graph on the matched entities for the purpose of data integration and other data discovery tasks. A recent approach [12] extends entity matching on unstructured text and tables by first building an entity knowledge graph and then creating node embeddings over the graph. The knowledge graph creation process, however, is largely dependent on external ontologies or knowledge bases which makes the solution less general. Entity matching discovery is a fundamentally different problem from the one addressed in this paper because the former works at the entity level, i.e., individual tuples (or even attributes within a tuple) in tables and entity mentions in documents, whereas our system works at the higher-level of columns, tables, and documents. The aggregation from the fine-grained entity level to the higher-order objects is an in interesting research problem in itself.

Other techniques rely on transforming structured tables into unstructured data, and then use a rich suite of language models to uncover relationships across tables and documents, e.g., [17, 19, 28]. For example, each tuple in a table is transformed into a sequence of tokens, and thus the entire table is transformed into a text file. This approach has been used for semantic matching and inductive reasoning [19]. Other systems such as Termite [32] moves the two modalities to a third common modality where everything is encoded as a set of subject-predicate-object triplets, e.g., [32]. Again, it loses the ability of supporting full-fledged discovery pipelines such as those highlighted in Section 1. Moreover, Termite uses these triplets for supervised learning of its model, whereas our system assumes no prior knowledge on the data lake and autonomously learns through weak supervision.

## 8 CONCLUSION

Data discovery, especially across modalities, is a core and yet challenging problem to modern enterprises. In this paper, we proposed CMDL as one step forward towards a holistic and end-to-end system that supports a superset of functionalities compared to the state-of-the-art data discovery systems. In particular, treating both tabular relations and unstructured document repositories as first-class citizens in the discovery process, seamlessly supporting discovery pipelines that intermixes tasks across the two domains, devising a novel embedding-based joint representation, and proposing a weakly-supervised framework for generating labeled data through a novel integration of CMDL's indexes. The evaluation results are promising and they show clear value-added benefits for the cross-modality discovery tasks, while at least retaining comparable performance to existing techniques for structured data discovery.